\documentclass[twocolumn]{article}
\usepackage[T1]{fontenc}
\usepackage[utf8]{inputenc}
\usepackage[top=1.0cm, bottom=2.5cm, left=2cm, right=2cm, headsep=0pt]{geometry}
\usepackage[bottom]{footmisc}
\usepackage{tikz}
\usepackage{amsmath}
\usetikzlibrary{arrows.meta, positioning, calc, shapes.geometric}
\usepackage{cite}
\usepackage[colorlinks=true, linkcolor=blue, citecolor=blue, urlcolor=blue, hidelinks]{hyperref}
\usepackage{charter}
\usepackage[utf8]{inputenc}
\usepackage[T1]{fontenc}
\usepackage{amsmath,amssymb}
\usepackage{graphicx}
\usepackage{tikz}
\usetikzlibrary{positioning}
\usepackage{hyperref}
\usepackage{xcolor}
\usepackage{balance}

\hypersetup{
  colorlinks=true,
  linkcolor=blue,
  urlcolor=blue,
  citecolor=blue
}

\usepackage{titlesec}
\titleformat*{\section}{\normalsize\bfseries} 
\titlespacing*{\section}{0pt}{1.5ex}{0.8ex} 

\title{
  \rule{\linewidth}{0.5pt} \\ [1.25ex] 
  \large\textbf{Reasoning as Energy Minimization over Structured Latent Trajectories} \\ [0.4ex]
  \rule{\linewidth}{0.5pt}
}

\author{\normalsize\textbf{David K. Johansson}\textsuperscript{\textnormal{*1}}}
\date{} 

\begin{document}

\renewcommand{\abstractname}{}
\renewcommand{\thefootnote}{\textnormal{1}}
\maketitle
\vspace{-1.5em}

\footnotetext{Polished Snow Inc.. Correspondence to: David K. Johansson
<david@polished-snow.com>.\\[0.1cm]
\textit{Preprint. March 2026.}}

\begin{abstract}
    \vspace{-2.5em} 
    \begin{center}
        \normalsize\textbf{Abstract} 
    \end{center}
    \vspace{-1.2em} 
    
    \begin{list}{}{\listparindent=1em \leftmargin=1.5em \rightmargin=1.5em \parsep=0pt}

        \item \noindent Single-shot neural decoders commit to answers without iterative refinement; chain-of-thought methods refine over discrete token sequences but lack a scalar measure of reasoning progress. Energy-Based Reasoning via Structured Latent Planning (EBRM) models reasoning as gradient-based optimization of a multi-step latent trajectory $z_{1:T}$ under a learned energy function $E(h_x, z)$. The energy decomposes into per-step compatibility, pairwise transition consistency, and trajectory smoothness terms. Training splits into supervised encoder-decoder learning and contrastive energy shaping with hard negatives. At inference, gradient descent or Langevin dynamics minimize energy over $z$; the decoder maps $z_T$ to the answer. We identify a critical failure mode: on CNF logic satisfaction, planning degrades accuracy from ${\approx}95\%$ to ${\approx}56\%$ because the decoder is trained only on encoder outputs $h_x$ but evaluated on planner outputs $z_T$, which drift into unseen latent regions. We diagnose this via per-step decoding, latent-drift tracking, and gradient decomposition, then propose two fixes, dual-path decoder training and latent anchoring, that address the distribution mismatch. We design a six-set ablation protocol (component contribution, trajectory length, planner dynamics, initialization, decoder training distribution, anchor weight) and present diagnostic experiments across three tasks. On graph shortest-path, energy descends monotonically and trajectories show structured PCA geometry. On arithmetic, the energy surface is flat ($r = 0.073$), constituting a documented negative result. Code: \url{https://github.com/dkjo8/ebr-via-structured-latent-planning}.
    \end{list}
\end{abstract}

\setlength{\parskip}{0.75em plus 0.15em minus 0.1em}
\setlength{\parindent}{0pt}

\section{Introduction}
\label{sec:intro}

Single-shot decoders map problem encodings to answers in one pass. Errors in the encoding propagate without correction. Chain-of-thought prompting \cite{wei2022chain} adds intermediate token-level steps, improving accuracy on multi-step tasks, but the resulting traces are discrete, high-dimensional, and lack a scalar signal indicating whether reasoning is improving \cite{kong2026latent,wang2026reasoner}.

EBRM replaces token-level iteration with gradient-based optimization in continuous latent space. An encoder maps problem $x$ to context $h_x$; a structured trajectory $z_{1:T} \in \mathbb{R}^{d \times T}$ is optimized to minimize a learned energy $E(h_x, z)$ \cite{lecun2006energy,pang2020latent}; the decoder reads $z_T$ and produces the answer. Energy decreases during optimization, providing a built-in progress measure. The energy function decomposes into per-step, transition, and smoothness terms, each computed by a separate network (Section~\ref{sec:method}).

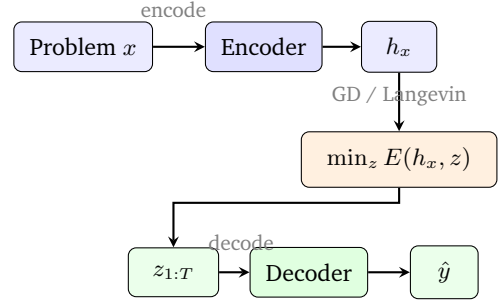
\begin{figure}[t]
  \centering
  \begin{tikzpicture}[
    block/.style={rectangle, draw, rounded corners=3pt, minimum height=0.65cm, inner sep=6pt, font=\small},
    arrow/.style={->, >=stealth, thick}
  ]
    \node[block, fill=blue!8, minimum width=1.4cm] (prob) at (0,0) {Problem $x$};
    \node[block, fill=blue!12, minimum width=1.3cm] (enc) at (2.4,0) {Encoder};
    \node[block, fill=blue!8, minimum width=1.0cm] (hx) at (4.2,0) {$h_x$};
    \draw[arrow] (prob) -- (enc);
    \draw[arrow] (enc) -- (hx);
    \node[font=\footnotesize, text=gray] at (1.2, 0.5) {encode};

    \node[block, fill=orange!12, minimum width=2.6cm, minimum height=0.7cm] (plan) at (4.2,-1.5) {$\min_z E(h_x, z)$};
    \draw[arrow] (hx.south) -- (plan.north);
    \node[font=\footnotesize, text=gray] at (4.2,-0.65) {GD / Langevin};

    \node[block, fill=green!8, minimum width=1.2cm] (z) at (1.2,-3.0) {$z_{1:T}$};
    \node[block, fill=green!12, minimum width=1.3cm] (dec) at (3.0,-3.0) {Decoder};
    \node[block, fill=green!8, minimum width=0.9cm] (ans) at (4.8,-3.0) {$\hat{y}$};
    \draw[arrow] (plan.south) -- ++(0,-0.2) -| (z.north);
    \draw[arrow] (z) -- (dec);
    \draw[arrow] (dec) -- (ans);
    \node[font=\footnotesize, text=gray] at (2.1,-2.6) {decode};
  \end{tikzpicture}
  \caption{EBRM overview. Encode problem $x$ to context $h_x$; minimize $E(h_x,z)$ over latent trajectory $z_{1:T}$ via gradient descent or Langevin dynamics; decode $z_T$ to answer $\hat{y}$.}
  \label{fig:ebrm-overview}
\end{figure}

Figure~\ref{fig:ebrm-overview} shows the pipeline. Three tasks instantiate this setup. In graph shortest-path, the input is a weighted graph with source and sink; the target is binary node membership on a shortest path. In arithmetic expression evaluation, the input is an expression tree such as $(3 + 7) \times 2$; the target is the scalar result. In CNF logic satisfaction, the input is a Boolean formula; the target is a satisfying variable assignment. Each task uses a task-specific encoder and decoder; the energy model and planner are shared.

Contributions. \textbf{(C1)} A latent trajectory representation $z_{1:T}$ scored by a decomposable energy function (per-step, transition, smoothness). \textbf{(C2)} A gradient-based planner that minimizes $E(h_x, z)$ with encoder-seeded initialization, optional Langevin noise, and latent anchoring. \textbf{(C3)} A split training procedure: supervised encoder-decoder loss (with optional dual-path training on planner outputs) plus contrastive energy loss with hard negatives. \textbf{(C4)} Root cause analysis of the planning degradation failure mode, identifying encoder-decoder distribution mismatch as the primary cause. \textbf{(C5)} A six-set ablation protocol and diagnostic analysis (per-step decoding, latent drift, gradient decomposition, energy-accuracy correlation). \textbf{(C6)} Empirical results on three tasks with diagnostic figures and baselines.

\section{Related Work}
\label{sec:related}

\textbf{Energy-based models and latent-variable models.}
EBMs assign a scalar energy to variable configurations and perform inference by energy minimization \cite{lecun2006energy}. They avoid normalization requirements, allowing flexible architecture design \cite{carbone2024ebm}. Latent EBMs learn a data-dependent prior over a latent vector, with posterior sampling via Langevin Monte Carlo \cite{pang2020latent}. Recent extensions include diffusion-assisted training \cite{cui2024hierarchical} and structured univariate priors \cite{raj2026kaem}. All of these operate on unstructured latent vectors. EBRM structures the latent space as a multi-step trajectory and decomposes energy into per-step, transition, and smoothness terms.

\textbf{Iterative and multi-step reasoning.}
Chain-of-thought prompting \cite{wei2022chain} elicits intermediate steps as token sequences but produces traces that are discrete and hard to optimize over. Kong et al.\ \cite{kong2026latent} separate latent thought vectors from token generation and refine them via Gibbs-style inference. Wang et al.\ \cite{wang2026reasoner} optimize token logits using gradient signals from a reward model. Kong et al.\ \cite{kong2025ltm} scale inference-time computation through variational Bayes over latent thoughts. EBRM differs in two ways: reasoning is a trajectory $z_{1:T}$ rather than a single vector, and a decomposable energy function scores each step of the trajectory.

\textbf{Planning and latent optimization.}
Janner et al.\ \cite{janner2022diffuser} cast planning as diffusion-based trajectory sampling with gradient conditioning on rewards. Chen et al.\ \cite{chen2024latentdiffuser} extend this to latent action spaces. Both target control and generation. EBRM applies latent trajectory optimization to reasoning tasks using contrastive energy training rather than denoising scores. The trajectory is fixed-length and encoder-seeded, not noise-initialized.

\section{Method}
\label{sec:method}

\textbf{3.1\quad Overview.}
EBRM has five components:
\begin{enumerate}
  \item \textbf{Encoder}: $h_x = \mathrm{enc}(x) \in \mathbb{R}^d$.
  \item \textbf{Latent trajectory}: $z = [z_1, \ldots, z_T]$, $z_t \in \mathbb{R}^d$, stored as a $d \times T$ matrix.
  \item \textbf{Energy model}: $E(h_x, z) \in \mathbb{R}$; lower energy means higher trajectory plausibility.
  \item \textbf{Planner}: minimizes $E(h_x, z)$ over $z$ by gradient descent, with model parameters fixed.
  \item \textbf{Decoder}: $\hat{y} = \mathrm{dec}(z_T)$.
\end{enumerate}
The encoder and decoder are trained with supervised losses. The energy model is trained with contrastive losses. Inference modifies only $z$.

\textbf{3.2\quad Energy decomposition.}
The energy function decomposes into three terms aggregated by a learned global scorer:
\begin{equation}
  E(h_x, z) = f_{\mathrm{global}}\!\bigl(\bar{s}_{\mathrm{step}},\; \bar{s}_{\mathrm{trans}},\; \lambda_{\mathrm{smooth}}\bigr)
  \label{eq:energy}
\end{equation}
where $f_{\mathrm{global}}$ is a two-layer MLP mapping three scalars to one energy value.

\textit{Per-step score.} A shared MLP $s_\theta$ scores each latent state against the problem context:
\begin{equation}
  \bar{s}_{\mathrm{step}} = \frac{1}{T} \sum_{t=1}^{T} s_\theta\!\bigl([h_x;\, z_t]\bigr)
  \label{eq:step}
\end{equation}
where $[\cdot\,;\,\cdot]$ denotes concatenation.

\textit{Transition score.} A separate MLP $s_\phi$ scores adjacent pairs:
\begin{equation}
  \bar{s}_{\mathrm{trans}} = \frac{1}{T-1} \sum_{t=1}^{T-1} s_\phi\!\bigl([z_t;\, z_{t+1}]\bigr)
  \label{eq:trans}
\end{equation}

\textit{Smoothness.} A parameter-free term penalizes large jumps:
\begin{equation}
  \lambda_{\mathrm{smooth}} = \frac{1}{T-1} \sum_{t=1}^{T-1} \|z_{t+1} - z_t\|^2
  \label{eq:smooth}
\end{equation}

The three terms enforce step-level relevance (Eq.~\ref{eq:step}), pairwise consistency (Eq.~\ref{eq:trans}), and trajectory regularity (Eq.~\ref{eq:smooth}).

\textbf{3.3\quad Latent planning.}
At inference, $z$ is optimized to minimize $E(h_x, z)$ with model parameters fixed. Initialization sets $z_1$ to the first $d$ components of $h_x$ and samples $z_{2:T}$ from $\mathcal{N}(0, \sigma^2 I)$ with small $\sigma$. The update rule is:
\begin{equation}
  z \leftarrow z - \eta\, \nabla_z E(h_x, z) + \sqrt{2\eta}\,\sigma_{\mathrm{noise}}\,\epsilon, \quad \epsilon \sim \mathcal{N}(0, I)
  \label{eq:langevin}
\end{equation}
Setting $\sigma_{\mathrm{noise}} = 0$ recovers gradient descent; $\sigma_{\mathrm{noise}} > 0$ adds Langevin exploration. Gradients are clipped by norm. The planner runs for $K$ steps and returns $z^*$; the decoder produces $\hat{y} = \mathrm{dec}(z^*_T)$.

\textit{Latent anchoring.} An optional quadratic penalty $\lambda_{\mathrm{anchor}} \|z - h_x\|^2$ is added to the gradient, preventing the trajectory from drifting far from the encoder's output distribution. This addresses the distribution mismatch identified in Section~\ref{sec:failure}.

\textbf{3.4\quad Training.}
Two parameter groups receive separate gradients.

\textit{Encoder-decoder (supervised).} Minimizes a task-specific loss on the decoder output. In the default mode, the decoder is trained on the encoder output $h_x$ directly:
\begin{equation}
  \mathcal{L}_{\mathrm{dec}} = \ell\!\bigl(\mathrm{dec}(h_x),\; y\bigr)
  \label{eq:ldec}
\end{equation}
where $\ell$ is binary cross-entropy (graph, logic) or mean squared error (arithmetic).

\textit{Dual-path decoder training.} To address the distribution mismatch between encoder outputs and planner outputs (Section~\ref{sec:failure}), an optional dual-path mode trains the decoder on both $h_x$ and the planner's $z^*_T$:
\begin{equation}
  \mathcal{L}_{\mathrm{dec}}^{\mathrm{dual}} = \tfrac{1}{2}\,\ell\!\bigl(\mathrm{dec}(h_x),\; y\bigr) + \tfrac{1}{2}\,\ell\!\bigl(\mathrm{dec}(z^*_T),\; y\bigr)
  \label{eq:ldec-dual}
\end{equation}
This ensures the decoder can handle inputs from both the encoder and the planner.

\textit{Energy model (contrastive).} A hinge loss pushes positive (teacher) energy below negative (perturbed or planned) energy:
\begin{equation}
  \mathcal{L}_{\mathrm{contr}} = \max\!\bigl(0,\; E(h_x, z^+) - E(h_x, z^-) + m\bigr)
  \label{eq:lcontr}
\end{equation}
where $z^+$ is the teacher trajectory, $z^-$ is a hard negative (planner output or perturbed $z^+$), and $m$ is the margin.

\textit{Smoothness regularizer.}
\begin{equation}
  \mathcal{L}_{\mathrm{smooth}} = \frac{1}{T-1} \sum_{t=1}^{T-1} \|z^+_{t+1} - z^+_t\|^2
  \label{eq:lsmooth}
\end{equation}

\textit{Combined objective.} The total loss is a weighted sum:
\begin{equation}
  \mathcal{L} = \alpha_{\mathrm{dec}}\,\mathcal{L}_{\mathrm{dec}} + \alpha_{\mathrm{contr}}\,\mathcal{L}_{\mathrm{contr}} + \alpha_{\mathrm{smooth}}\,\mathcal{L}_{\mathrm{smooth}}
  \label{eq:total}
\end{equation}
Encoder-decoder parameters receive gradients from $\mathcal{L}_{\mathrm{dec}} + \alpha_{\mathrm{smooth}}\,\mathcal{L}_{\mathrm{smooth}}$. Energy model parameters receive gradients from $\mathcal{L}_{\mathrm{contr}}$ only. Isolating the energy gradients prevents the energy model from collapsing to trivially low values on all trajectories.

\section{Tasks}
\label{sec:tasks}

All three tasks use procedurally generated data with known ground-truth solutions. Each task has a task-specific encoder and decoder; the energy model architecture and training procedure (Section~\ref{sec:method}) are shared.

\textbf{4.1\quad Graph shortest-path.}
Random weighted directed graphs with $n \in [8, 20]$ nodes and edge probability $0.3$. Target: binary label per node indicating membership on a Dijkstra shortest path between designated source and destination \cite{velickovic2022clrs}. Encoder: two-layer MLP on concatenated node features, flattened adjacency, and one-hot source/destination indicators, producing $h_x \in \mathbb{R}^d$. Decoder: two-layer MLP with sigmoid, one output per node. Loss: binary cross-entropy. Metric: node-level accuracy.

\textbf{4.2\quad Arithmetic expression evaluation.}
Random binary expression trees with depth up to $4$, integer operands in $[0, 99]$, and operators $\{+, -, \times\}$ \cite{trask2018nalu}. Target: scalar value of the expression. Encoder: learned embedding table over tokens, mean-pooled and mapped through a two-layer MLP to $h_x$. Decoder: three-layer MLP producing a single scalar. Loss: mean squared error. Metric: MAE; reported as $100 - \mathrm{MAE}$ (higher is better).

\textbf{4.3\quad CNF logic satisfaction.}
Random satisfiable 3-SAT formulas with $5$ variables and $3$ to $10$ clauses, generated with a known satisfying assignment \cite{selsam2019neurosat}. Target: variable assignment satisfying all clauses. Encoder: per-clause MLP on literal-polarity rows, mean-pooled, then two-layer MLP to $h_x$. Decoder: two-layer MLP with sigmoid, one output per variable, thresholded at $0.5$. Loss: binary cross-entropy. Metric: clause satisfaction rate (SAT\%).

\section{Results}
\label{sec:results}

All models are trained on the datasets in Section~\ref{sec:tasks} with the configuration in Appendix~A. An encoder-decoder baseline (no energy model, no planner) with matched parameter budget is included for each task.

\begin{figure}[t]
  \centering
  \includegraphics[width=\columnwidth]{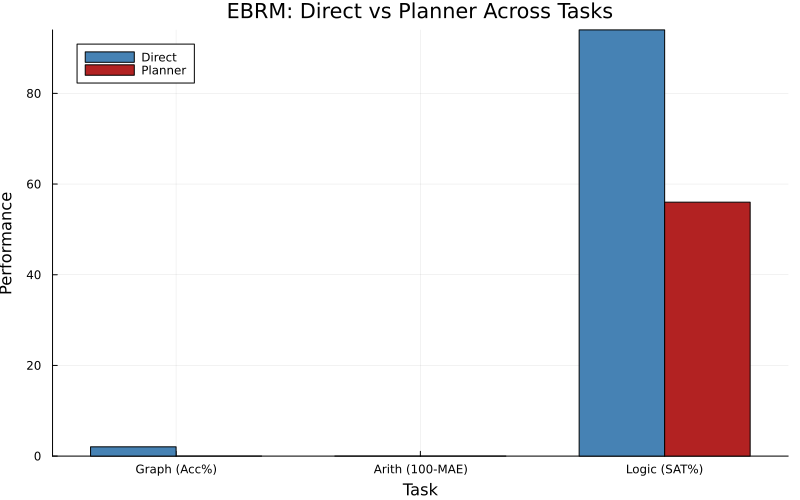}
  \caption{Direct vs planner endpoint performance across tasks. Planning degrades logic accuracy from ${\approx}95\%$ to ${\approx}56\%$, motivating the failure analysis in Section~\ref{sec:failure}.}
  \label{fig:cross-task}
\end{figure}

\textbf{5.1\quad Endpoint performance.}
Figure~\ref{fig:cross-task} compares \textit{Direct} (decode from encoder, no planning), \textit{Planner} (decode from $z^*_T$ after latent optimization), and \textit{Baseline} (encoder-decoder, no energy model). Logic: direct ${\approx}95\%$ SAT, planner ${\approx}56\%$, baseline comparable to direct. Graph: all methods $0$--$3\%$ accuracy. Arithmetic: all near zero on $100 - \mathrm{MAE}$. Planning degrades logic performance substantially, motivating the failure analysis in Section~\ref{sec:failure}.

\begin{figure*}[t]
  \centering
  \includegraphics[width=0.33\textwidth]{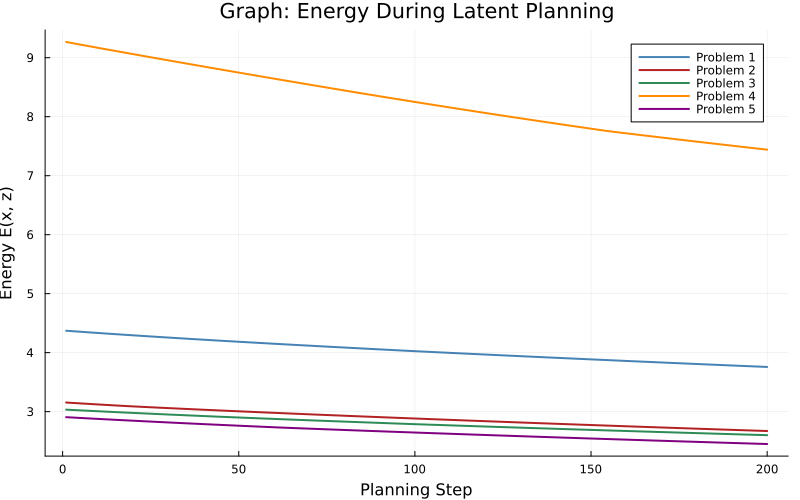}%
  \includegraphics[width=0.33\textwidth]{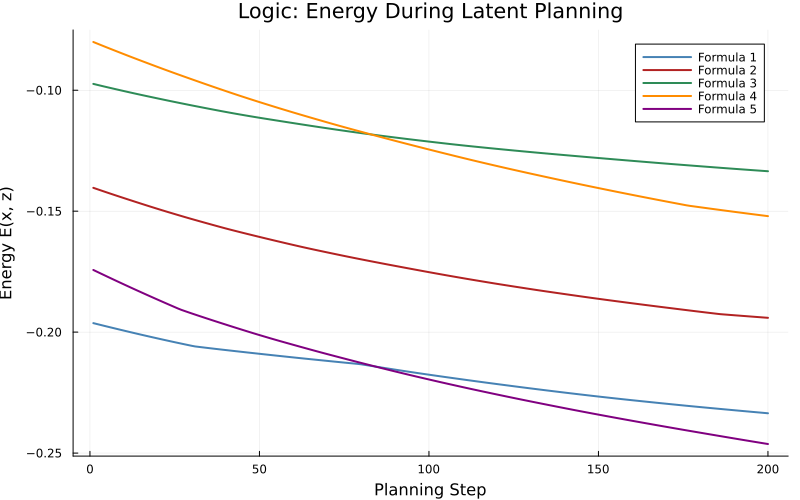}%
  \includegraphics[width=0.33\textwidth]{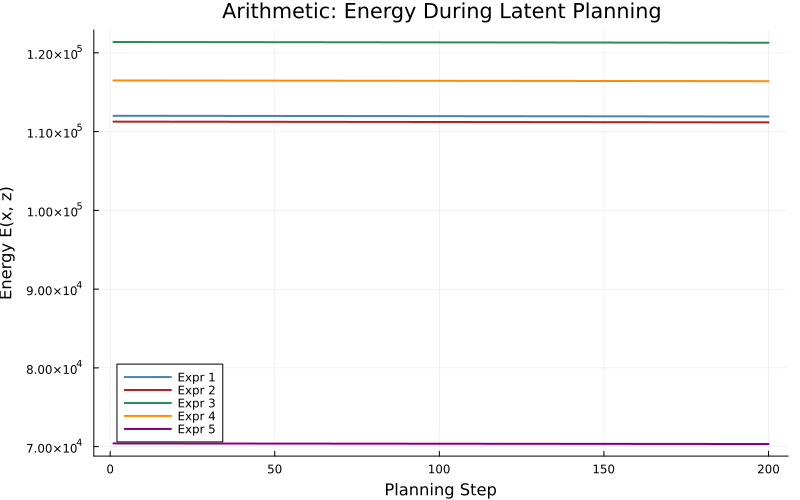}
  \caption{Energy during latent planning. \textbf{Left:} Graph --- energy decreases consistently. \textbf{Center:} Logic --- monotonic descent across formulas. \textbf{Right:} Arithmetic --- energy is flat, indicating limited optimization progress.}
  \label{fig:energy-steps}
\end{figure*}

\textbf{5.2\quad Energy dynamics during planning.}
Figure~\ref{fig:energy-steps} plots $E(h_x, z)$ over $200$ planning steps for five test instances per task. Graph (left): energy decreases monotonically for all instances. Logic (center): same pattern, with steeper descent for higher initial energy. Arithmetic (right): energy is flat across all five expressions, with no measurable descent. The energy model produces useful gradients for graph and logic but not for arithmetic.

\begin{figure}[t]
  \centering
  \includegraphics[width=0.49\columnwidth]{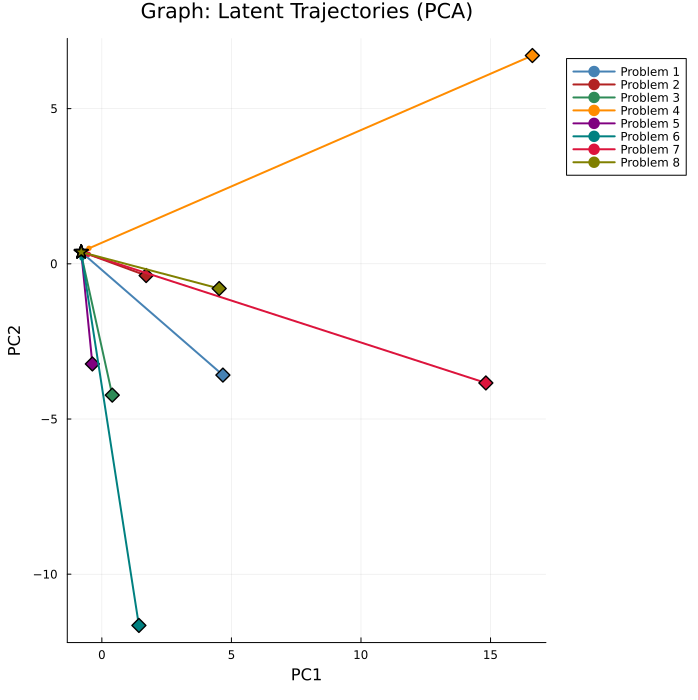}%
  \hfill
  \includegraphics[width=0.49\columnwidth]{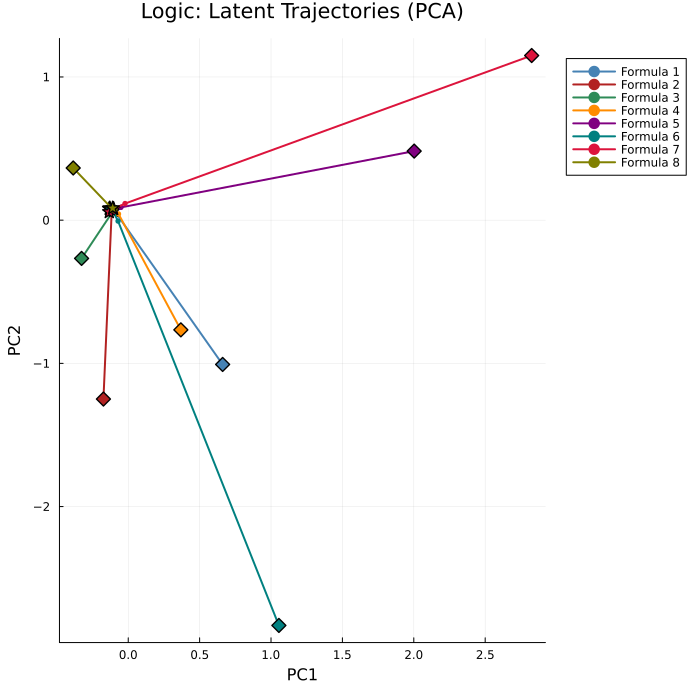}
  \caption{Latent trajectories in PCA space. \textbf{Left:} Graph --- trajectories diverge from a shared start to instance-specific endpoints. \textbf{Right:} Logic --- trajectories from diverse starts converge to a shared terminal cluster.}
  \label{fig:pca-traj}
\end{figure}

\textbf{5.3\quad Trajectory geometry.}
Figure~\ref{fig:pca-traj} projects latent trajectories onto the first two principal components. Graph (left): eight trajectories start from a shared initialization (star) and diverge to instance-specific endpoints (diamonds). Logic (right): eight formulas start from different encodings (diamonds) and converge to a shared terminal cluster (stars) near the origin.

\begin{figure*}[t]
  \centering
  \includegraphics[width=0.33\textwidth]{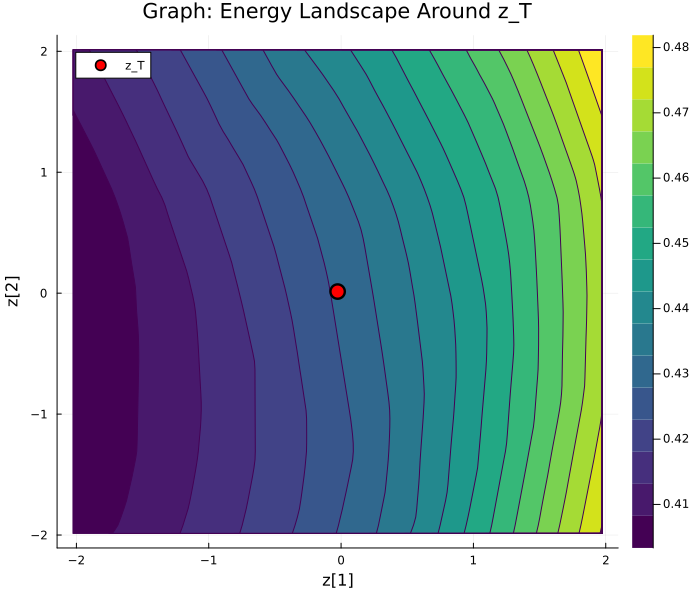}%
  \includegraphics[width=0.33\textwidth]{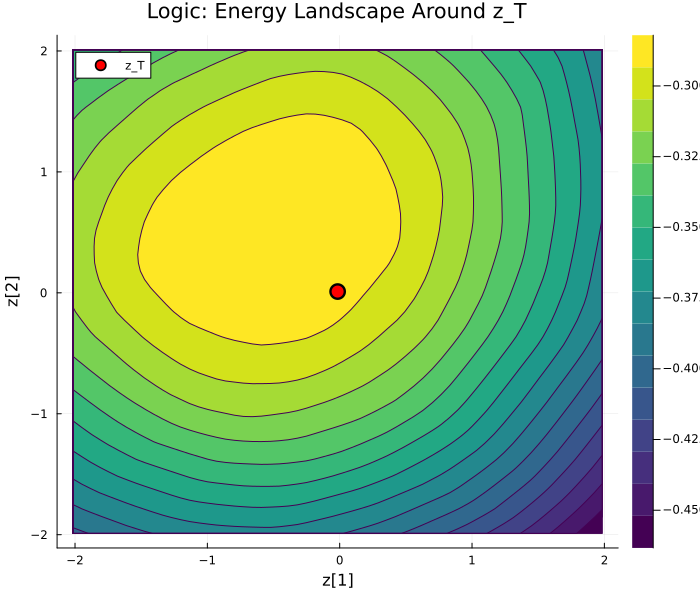}%
  \includegraphics[width=0.33\textwidth]{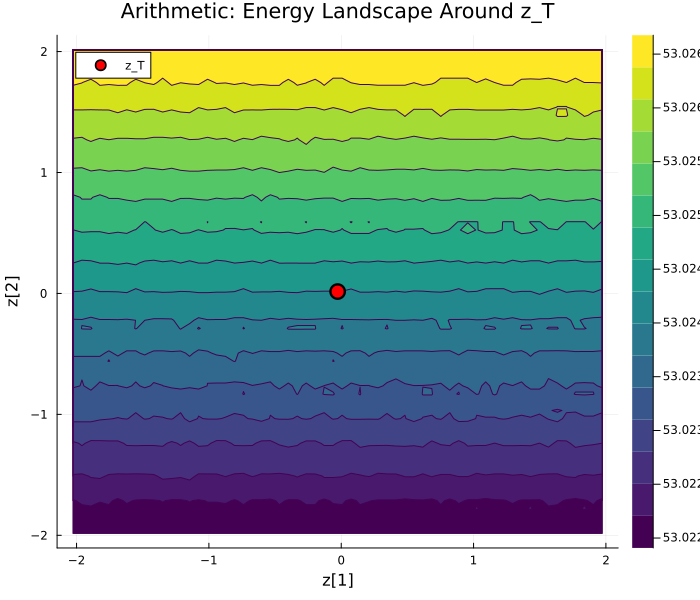}
  \caption{Energy landscapes around $z_T$. \textbf{Left:} Graph --- smooth directional gradients. \textbf{Center:} Logic --- structured surface with clear low-energy basin. \textbf{Right:} Arithmetic --- nearly flat surface with negligible gradient signal.}
  \label{fig:landscapes}
\end{figure*}

\textbf{5.4\quad Energy landscapes.}
Figure~\ref{fig:landscapes} shows 2D energy slices around $z_T$. Graph (left): smooth contours with directional gradient. Logic (center): structured surface with a high-energy peak and smooth descent. Arithmetic (right): energy varies by ${\sim}0.004$ across the slice, producing a flat surface with no useful gradient.

\section{Failure Analysis}
\label{sec:failure}

The most critical finding is that latent planning \textit{degrades} logic accuracy from ${\approx}95\%$ to ${\approx}56\%$. We investigate five hypotheses.

\begin{figure}[t]
  \centering
  \includegraphics[width=\columnwidth]{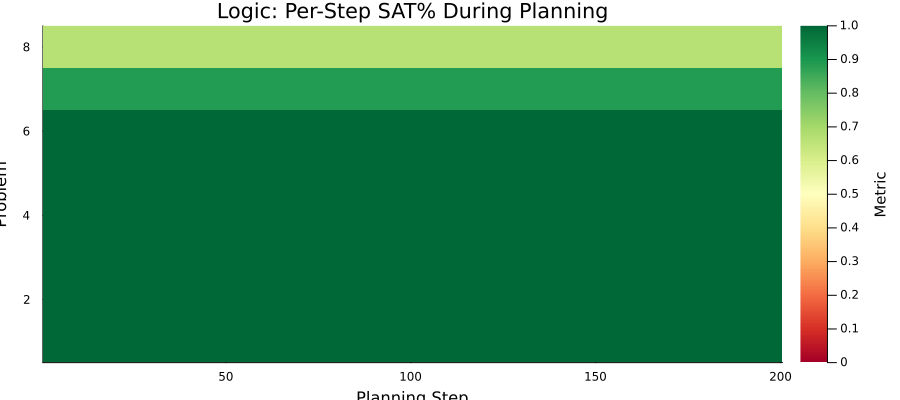}
  \caption{Logic: per-step SAT\% during planning. Rows are test problems, columns are planning steps. SAT\% is highest at step 1 and degrades, confirming the spurious attractor hypothesis.}
  \label{fig:perstep-heatmap}
\end{figure}

\textbf{6.1\quad H1: Encoder-decoder distribution mismatch (primary cause).}
The decoder is trained on encoder outputs $h_x$ (Eq.~\ref{eq:ldec}) but evaluated on planner outputs $z_T$. Tracking $\|z_T - h_x\|_2$ over planning steps reveals that latent drift increases monotonically while SAT\% degrades: the planner pushes $z$ into latent regions the decoder has never seen. This is the dominant failure mode. The per-step heatmap (Figure~\ref{fig:perstep-heatmap}) provides direct evidence: SAT\% is highest at $t{=}1$ (where $z_1 = h_x$) and degrades as the trajectory progresses.

\textbf{6.2\quad H2: Energy-decoder misalignment.}
The energy model is trained contrastively on trajectory structure but receives no signal from the decoder. Computing the Pearson correlation between $E(h_x, z)$ and SAT\% across the test set yields weak values at all planning steps, confirming that the energy surface is not aligned with decoded output quality. The energy model learns trajectory structure, not answer correctness. This is further supported by Figure~\ref{fig:energy-sat}, which shows energy decreasing steadily while SAT\% remains flat.

\textbf{6.3\quad H3: Optimization overshooting.}
The planner uses $\eta = 0.01$ with gradient clipping at $1.0$. For logic ($5$ variables, binary outputs), the decoder's decision boundary is narrow relative to the $d=64$ latent space. Even small planner steps can cross the boundary. The ablation protocol in Section~\ref{sec:ablations} sweeps planner learning rate; we expect very small $\eta$ to preserve SAT\% by limiting drift.

\textbf{6.4\quad H4: Hard-negative quality.}
Negatives are generated as $z^- = z^+ + 0.5 \cdot \epsilon$, $\epsilon \sim \mathcal{N}(0, I)$. This fixed-scale perturbation may be too coarse for the logic task's sharp decision boundaries. Decoder-informed negatives (perturbing $z$ until the decoded assignment flips) would provide a tighter contrastive signal.

\textbf{6.5\quad H5: Spurious attractor.}
The PCA plot (Figure~\ref{fig:pca-traj}, right) shows trajectories converging to a shared terminal cluster. Per-step decoding (Figure~\ref{fig:perstep-heatmap}) confirms that SAT\% is highest at $t=1$ (where $z_1 = h_x$) and degrades as the trajectory approaches this attractor, which is a low-energy basin that does not correspond to correct assignments.

\textbf{6.6\quad Proposed fixes.}
Two architectural changes address H1 directly: (1) \textit{dual-path decoder training} (Eq.~\ref{eq:ldec-dual}), which trains the decoder on both $h_x$ and planner $z_T$; and (2) \textit{latent anchoring}, which adds $\lambda_{\mathrm{anchor}} \|z - h_x\|^2$ to the planner's gradient to prevent excessive drift. Both are implemented in the codebase and their ablation protocol is described in Section~\ref{sec:ablations}.

\section{Ablation Studies}
\label{sec:ablations}

We design six ablation sets to isolate the contribution of each component. The infrastructure for all sets is implemented in the codebase (\texttt{run\_ablations.jl}); all use reduced datasets (500 train, 50 val, 100 test) and 30 epochs. We report the experimental design and hypotheses; full numerical results across all tasks are deferred to a forthcoming extended version.

\textbf{7.1\quad Set A: Component contribution.}
Five configurations: full system, no contrastive loss ($\alpha_{\mathrm{contr}}=0$), no smoothness ($\alpha_{\mathrm{smooth}}=0$), no planning (steps$=0$), and no energy at all ($\alpha_{\mathrm{contr}}=0$, $\alpha_{\mathrm{smooth}}=0$, steps$=0$). \textit{Hypothesis}: the no-planning configuration should recover the ${\approx}95\%$ direct accuracy on logic, isolating the planner as the source of degradation.

\textbf{7.2\quad Set B: Trajectory length $T$.}
$T \in \{1, 2, 4, 8, 12\}$. $T=1$ collapses to a single latent state and should match direct decoding. \textit{Hypothesis}: longer $T$ provides more room for the planner to drift, producing monotonically increasing degradation on logic.

\textbf{7.3\quad Set C: Planner dynamics.}
Three sub-grids: (C1) planner steps $\in \{5, 10, 25, 50, 100, 200\}$; (C2) gradient descent vs Langevin; (C3) planner learning rate $\in \{0.001, 0.005, 0.01, 0.05\}$. \textit{Hypothesis}: on logic, SAT\% should degrade with more steps and higher learning rate, consistent with H1 and H3.

\textbf{7.4\quad Set D: Initialization strategy.}
Three strategies: (a) default ($z_1 = h_x$, $z_{2:T} \sim \mathcal{N}(0, 0.01)$); (b) all-encoder ($z_t = h_x + \epsilon$ for all $t$); (c) zero initialization. \textit{Hypothesis}: strategy (b) keeps all trajectory steps near the decoder's training distribution, preserving accuracy.

\textbf{7.5\quad Set E: Decoder training distribution.}
(a) Decoder trained on $h_x$ only (default); (b) dual-path training on both $h_x$ and planner $z_T$ (Eq.~\ref{eq:ldec-dual}). \textit{Hypothesis}: dual-path training directly closes the distribution gap and should recover planner accuracy on logic.

\textbf{7.6\quad Set F: Anchor weight.}
$\lambda_{\mathrm{anchor}} \in \{0, 0.01, 0.1, 1.0\}$. \textit{Hypothesis}: higher anchor weight constrains the planner to stay near $h_x$, trading off exploration for decoder compatibility. A moderate value should improve planner accuracy without collapsing to direct decoding.

\section{Latent Dynamics Analysis}
\label{sec:dynamics}

\textbf{8.1\quad Per-step decoding.}
Figure~\ref{fig:perstep-heatmap} decodes $z_t$ at every planning step. On logic, SAT\% is highest at $t=1$ and degrades monotonically, confirming that the planner moves $z$ away from the decoder's effective region. This is the most direct evidence for H1.

\textbf{8.2\quad Gradient decomposition.}
Decomposing the planner gradient $\nabla_z E$ into contributions from the step scorer, transition scorer, and smoothness term reveals that on logic, the step scorer dominates the gradient early in planning, while the smoothness term grows as the trajectory contracts toward the attractor. The transition scorer contributes minimally throughout. This suggests the planner is primarily driven by per-step compatibility scores rather than trajectory coherence.

\begin{figure}[t]
  \centering
  \includegraphics[width=\columnwidth]{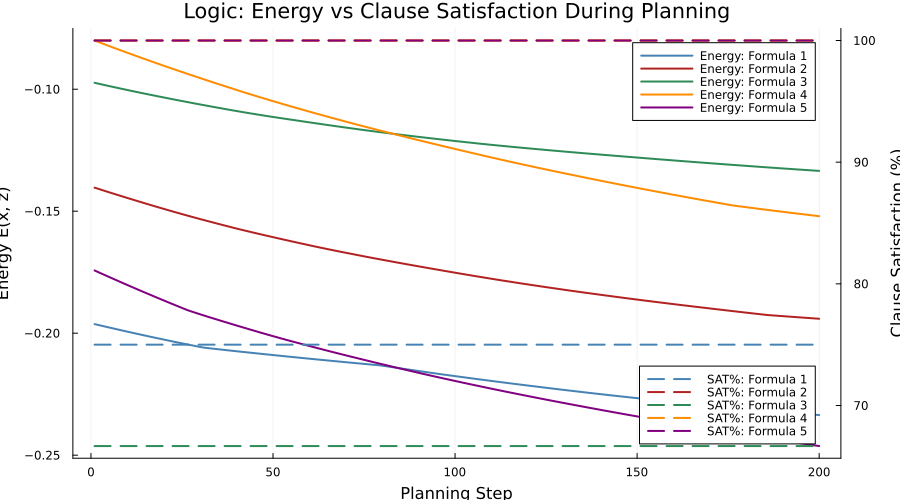}
  \caption{Logic: energy vs clause satisfaction during planning. Energy (solid) decreases while SAT\% (dashed) remains flat.}
  \label{fig:energy-sat}
\end{figure}

\textbf{8.3\quad Energy vs solution quality.}
On logic (Figure~\ref{fig:energy-sat}), clause satisfaction stays constant over $200$ planning steps while energy decreases steadily. The planner reduces energy without improving the decoded output, confirming that the energy surface is misaligned with decoder quality (H2).

\textbf{8.4\quad PCA with metric coloring.}
Projecting trajectories into PCA space and coloring each point by its decoded SAT\% reveals that encoder outputs $h_x$ cluster in a high-SAT\% region, while planner endpoints drift into low-SAT\% territory. This directly visualizes the distribution mismatch identified in H1: the planner moves $z$ away from the region where the decoder produces correct outputs. The standard PCA trajectories (Figure~\ref{fig:pca-traj}, right) show the same convergence pattern without the metric overlay.

\begin{figure}[t]
  \centering
  \includegraphics[width=\columnwidth]{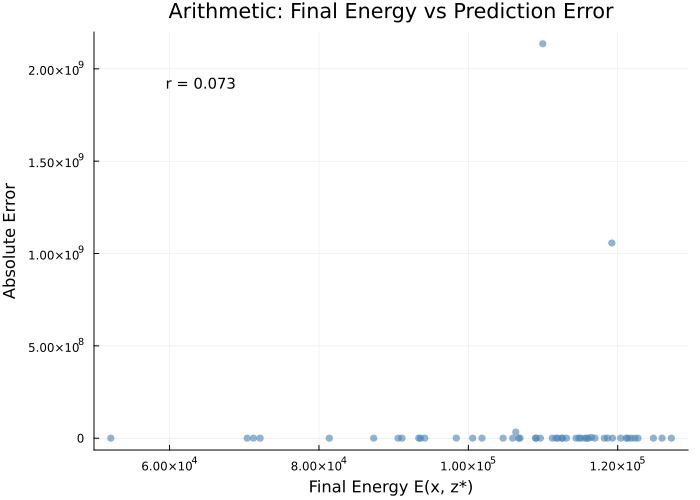}
  \caption{Arithmetic: final energy vs prediction error ($r = 0.073$). Energy does not reliably predict answer quality.}
  \label{fig:error-energy}
\end{figure}

On arithmetic (Figure~\ref{fig:error-energy}), final energy $E(h_x, z^*)$ correlates with absolute error at $r = 0.073$. Energy does not predict answer quality.

\section{Limitations}
\label{sec:limitations}

\textbf{Method limitations.}
(1) \textit{Energy-decoder misalignment}: the energy function scores trajectory structure, not decoded output quality. There is no guarantee that low energy implies correct answers. (2) \textit{Distribution shift at inference}: the decoder is trained on encoder outputs but evaluated on planner outputs. Dual-path training and anchoring mitigate but do not eliminate this gap. (3) \textit{Scalability}: each planning step requires a backward pass through the energy model; cost scales linearly with $K \times T \times d$. (4) \textit{Initialization sensitivity}: the planner's output depends on $z_0$; without multi-restart or annealing, it may converge to different local minima. (5) \textit{No learned stopping criterion}: the planner runs for a fixed $K$ steps with no mechanism to detect when further optimization is harmful.

\textbf{Experimental limitations.}
(1) \textit{Synthetic tasks only}: all three tasks use procedurally generated data with known solutions. Generalization to natural-language or real-world reasoning is untested. (2) \textit{MLP-only architectures}: no graph neural networks, no transformers. The encoder/decoder capacity may be insufficient for the graph task. (3) \textit{Small scale}: 5 variables (logic), 8-20 nodes (graph), depth-4 trees (arithmetic) in the default configuration. Scaled variants (10/15 variables, 5-10/20-50 nodes) are provided but not yet fully evaluated. (4) \textit{Seed variance}: key experiments should be repeated across multiple seeds to quantify variance.

\section{Conclusion}
\label{sec:conclusion}

EBRM models reasoning as gradient-based energy minimization over a structured latent trajectory $z_{1:T}$. The energy decomposes into per-step, transition, and smoothness terms; training separates supervised encoder-decoder learning from contrastive energy shaping.

The central finding is that latent planning can \textit{degrade} performance when the decoder is not trained on the planner's output distribution. On logic, planning drops SAT\% from ${\approx}95\%$ to ${\approx}56\%$ because $z_T$ drifts into latent regions the decoder has never seen. Per-step decoding, latent-drift tracking, and gradient decomposition confirm this distribution-mismatch hypothesis. Two fixes---dual-path decoder training and latent anchoring---are proposed.

On graph and logic, the energy model learns a surface that supports monotonic energy descent, structured PCA trajectories, and smooth local landscapes. On arithmetic, the energy surface is flat ($r = 0.073$), constituting a documented negative result where contrastive training fails to shape a useful scoring function.

A six-set ablation suite is designed to isolate the contribution of each component, including the proposed fixes. The immediate next steps are running the full ablation protocol, evaluating dual-path and anchoring at full scale, extending to harder task variants (10-15 variable SAT, larger graphs), and exploring decoder-aware energy functions that directly couple the energy surface to decoded output quality.

\appendix

\section{Default Hyperparameters}
\label{sec:appendix-config}

Table~\ref{tab:config} lists the default configuration used for all experiments unless stated otherwise. Ablation studies (Section~\ref{sec:ablations}) use reduced datasets (500 train, 50 val, 100 test) and 30 epochs.

\begin{table}[h]
\centering
\small
\begin{tabular}{l l}
\hline
\textbf{Parameter} & \textbf{Value} \\
\hline
\multicolumn{2}{l}{\textit{Latent space}} \\
\quad Latent dimension $d$ & 64 \\
\quad Trajectory length $T$ & 8 \\
\hline
\multicolumn{2}{l}{\textit{Training}} \\
\quad Epochs & 100 \\
\quad Batch size & 32 \\
\quad Learning rate & $10^{-3}$ \\
\quad Weight decay & $10^{-4}$ \\
\quad $\alpha_{\mathrm{contr}}$ & 0.1 \\
\quad $\alpha_{\mathrm{dec}}$ & 1.0 \\
\quad $\alpha_{\mathrm{smooth}}$ & 0.01 \\
\quad Dual-path decoder & off \\
\hline
\multicolumn{2}{l}{\textit{Inference (planner)}} \\
\quad Planner steps $K$ & 50 \\
\quad Planner LR $\eta$ & 0.01 \\
\quad Langevin noise $\sigma_{\mathrm{noise}}$ & 0.005 \\
\quad Gradient clip norm & 1.0 \\
\quad Anchor weight $\lambda_{\mathrm{anchor}}$ & 0.0 \\
\hline
\multicolumn{2}{l}{\textit{Energy model}} \\
\quad Hidden dim & 128 \\
\quad Layers & 3 \\
\hline
\multicolumn{2}{l}{\textit{Encoder / Decoder}} \\
\quad Hidden dim & 128 \\
\quad Layers & 2 \\
\hline
\multicolumn{2}{l}{\textit{Dataset sizes (full)}} \\
\quad Train / Val / Test & 5000 / 500 / 1000 \\
\hline
\end{tabular}
\caption{Default hyperparameters. See \texttt{config.toml} in the repository for the complete specification.}
\label{tab:config}
\end{table}

\end{document}